\documentclass[10pt, twocolumn]{article} %

\usepackage[american]{babel}

\usepackage{natbib} %
\usepackage{mathtools} %
\usepackage{booktabs} %

\usepackage{tikz}
\usepackage{tikz-network}
\usepackage{calc}       %
\usetikzlibrary{
   arrows,
   shapes,
   positioning,
   fit,
   backgrounds,
   chains,
   decorations.pathreplacing
}

\usepackage{algorithm}
\usepackage{fancyhdr}
\fancypagestyle{firstpage}{
  \fancyhf{}
  \fancyfoot[C]{Accepted for publication at Uncertainty in Artificial Intelligence (UAI) 2025.}

}
\usepackage{algorithmicx}
\usepackage{algpseudocode}
\usepackage{amsmath}
\usepackage{amssymb}
\usepackage{booktabs}
\usepackage{graphicx}
\usepackage{multirow}
\usepackage{multicol}
\usepackage{tabularray}
\usepackage{changepage} 
\usepackage{caption}
\usepackage{subcaption}
\usepackage{tcolorbox}
\usepackage{listings}
\usepackage{xcolor}
\usepackage{dsfont}
\usepackage{mdframed}

\usepackage[T1]{fontenc}
\usepackage{textcomp}
\usepackage{times}
\usepackage[scaled]{helvet}
\usepackage{courier}
\usepackage{microtype}

\usepackage[
    a4paper,
    paperheight=11in,
    textheight=9.25in,
    textwidth=6.75in,
    columnsep=0.25in,
    nohead,
    centering
]{geometry}

\newcommand{\methodname}[0]{\textsc{GPS}}

\newcommand{\ceil}[1]{\left\lceil #1 \right\rceil}

\newtheorem{theorem}{Theorem}[section]

\newtheorem{proposition}[theorem]{Proposition}

\algrenewcommand\algorithmicrequire{\textbf{Input:}}
\algrenewcommand\algorithmicensure{\textbf{Output:}}

\title{Conformal Prediction Sets for Deep Generative Models \\ via Reduction to Conformal Regression}
\author{
    \textbf{Hooman Shahrokhi}\thanks{These authors contributed equally to this work}\label{equalcontrib} \quad
    \textbf{Devjeet Raj Roy}\textsuperscript{*} \\
    \textbf{Yan Yan}\quad
    \textbf{Venera Arnaoudova}\quad
    \textbf{Janardhan Rao Doppa} \\ [1ex]
    School of EECS, Washington State University\\[0.5ex]
    \begin{tabular}{c}
    \texttt{\{hooman.shahrokhi, devjeet.roy, yan.yan,}\\
    \texttt{venera.arnaoudova, jana.doppa\}@wsu.edu}
    \end{tabular}
    \\
}
  \date{}

\begin{document}
\maketitle
\thispagestyle{firstpage}
\begin{abstract}
 We consider the problem of generating valid and small prediction sets by sampling outputs (e.g., software code and natural language text) from a black-box deep generative model for a given input (e.g., textual prompt). 
The validity of a prediction set is determined by a user-defined binary admissibility function depending on the target application. For example, requiring at least one program in the set to pass all test cases in code generation application.
To address this problem, we develop a simple and effective conformal inference algorithm referred to as {\em Generative Prediction Sets (GPS)}. Given a set of calibration examples and black-box access to a deep generative model, GPS can generate prediction sets with provable guarantees. The key insight behind GPS is to exploit the inherent structure within the distribution over the minimum number of samples needed to obtain an admissible output to develop a simple conformal regression approach over the minimum number of samples. %
Experiments on multiple datasets for code and math word problems using different large language models demonstrate the efficacy of GPS over state-of-the-art methods.

\end{abstract}

\section{Introduction}

Conditional deep generative models (DGMs) have recently become the dominant paradigm in a wide range of machine learning problems arising in domains including natural language processing, computer vision, and software engineering. Notable examples of DGMs include large language models (LLMs)~(\cite{brown2020language}), diffusion models~(\cite{yang2023diffusion}), and vision transformers~(\cite{khan2022transformers}). While these DGMs have displayed increasingly powerful predictive performance across a variety of generative tasks, they are far from infallible. For example, LLMs are known to produce hallucinations~(\cite{rawte2023survey}). %
Such erroneous outputs 
pose challenges in their use in safety-critical domains including medicine, law, education, and finance. For example, in many modern IDEs, users are able to query a LLM to generate programs~ \cite{asare2023github}. It would be desirable to attach to this set of suggested outputs a measure of uncertainty.  This motivates the need for theoretically-sound uncertainty quantification for deep generative models to deploy them safely.

\begin{figure*}
\centering
\includegraphics[width=\textwidth]{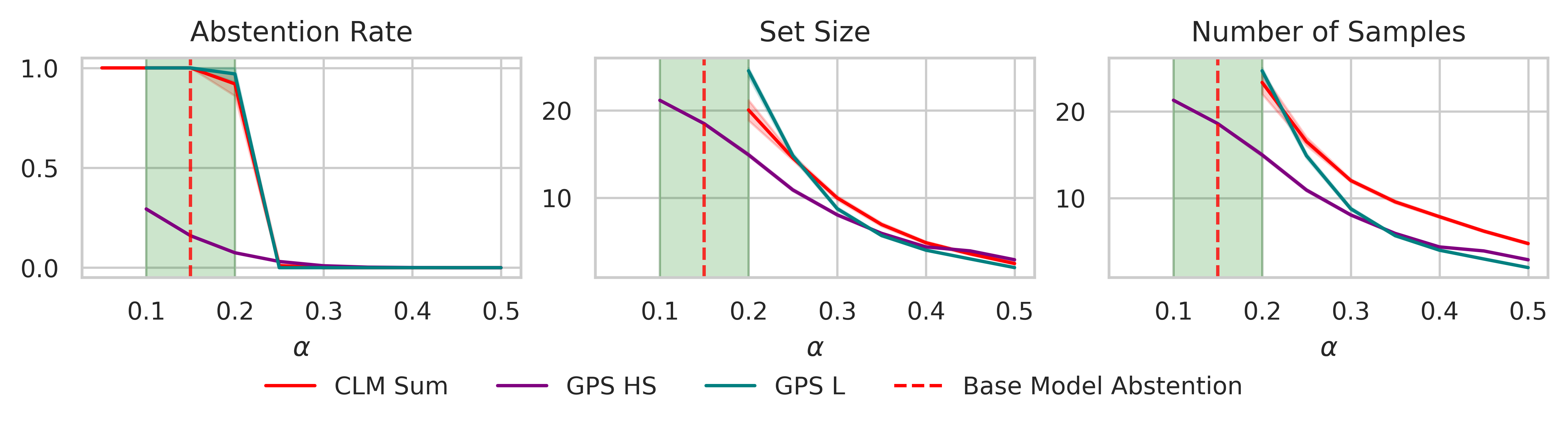}
\caption{Abstention rate, set sizes, and number of samples collected from the GPT-4o-mini on the MATH benchmark for CLM and two variants of GPS. %
Since CLM provides PAC-style guarantees, we adjust the $\alpha$ level of GPS to equate the coverage guarantee. GPS HS produces valid prediction sets over tight $\alpha$ levels where CLM does not (green shaded area).} %

\label{fig:gps-motivating-example}
\end{figure*}
Conformal prediction (CP)~(\cite{Vovk2005-cp,lei2014distribution}) is a general uncertainty quantification framework to produce prediction sets with theoretical guarantees for any black-box predictive model. Given some calibration data and a user-specified significance level $\alpha$, CP generates sets which are guaranteed to contain the correct output with probability $1-\alpha$ (referred to as {\em coverage}) through a calibration step. However, in the absence of an ordering over outputs, CP requires an enumeration of the entire output space, making it difficult to apply to deep generative models with combinatorial output spaces.

There is little work on this challenging problem setting. Conformal Language Modeling (CLM) \cite{Quach2023-mq}, constructs prediction sets iteratively by examining each sampled output. It calibrates three parameters $(\lambda_1, \lambda_2, \lambda_3)$: $\lambda_1$ determines if enough high-quality samples have been collected while $\lambda_2$ and $\lambda_3$ filter the collected samples based on quality criteria (e.g. total log probability of the samples). CLM uses Learn-Then-Test (\citet{angelopoulos2021learn}) to find valid configurations of these parameters that provide coverage at the user-specified confidence level $\alpha$ with a  high probability. However, it is not always possible to find valid configurations for tight confidence levels. Any calibration algorithm operating with finite sampling is fundamentally constrained by the base model's capacity to generate correct outputs within this finite horizon. To maintain distribution-free coverage when users request prediction sets at $\alpha$ levels exceeding the base model's abstention rate, we must output the entire space $\mathcal{Y}$ for at least some inputs (referred to as {\em abstention}). We can see this phenomenon in Figure~\ref{fig:gps-motivating-example}.

On the MATH benchmark, GPT-4o-mini produces a correct output within the first 25 samples approximately 85\% of the time (marginally on the joint distribution). The dashed red line shows this base abstention rate. In practice, CLM's performance is even more constrained -- it abstains for 100\% of our data until $\alpha = 0.2$, and does not achieve a 0\% abstention rate until $\alpha=0.25$. This gap emerges because CLM's filtering parameters can incorrectly reject correct solutions with non-zero probability, effectively inflating the model's abstention rate. Moreover, CLM's abstention is binary: for a given calibration dataset, CLM either produces a valid configuration, in which case it \textit{never} abstains, or fails to produce a valid configuration, in which it \textit{always} abstains. As a consequence of these limitations, 1) we need to collect more samples to produce valid prediction sets, and 2) if $\alpha$ is close enough to the given DGM's abstention rate, the calibration algorithm must abstain with non-zero probability.

In this work, we present a simple conformal calibration algorithm, referred to as {\em Generative Prediction Sets (GPS)}, to address these limitations. \methodname\ reduces the problem of constructing prediction sets for DGMs to a conformal regression problem: we learn an auxiliary predictor to estimate how many samples the model needs to produce a correct output, and employ conformal prediction (CP) tools~\cite{romano2019conformalized} to obtain valid prediction intervals around these estimates. This interval determines the number of samples we collect at test time to construct prediction sets in the original output space. A basic instantiation of \methodname\ (\texttt{GPS L} in Figure~\ref{fig:gps-motivating-example}) uses only the prompt's log probability as input features and can achieve similar abstention rates, set sizes, and sample efficiency as CLM without examining the sampled outputs by calibrating just a single parameter using vanilla CP. Thus, it is both simpler to implement in practice and more computationally efficient. If we incorporate richer signals from the underlying DGM (e.g., hidden state activations for the input prompt), the resulting variant (\texttt{GPS HS} in Figure~\ref{fig:gps-motivating-example}) can achieve dramatically lower abstention rates and valid prediction sets at $\alpha$ levels closer to the base model's true abstention rate, while maintaining set sizes competitive with CLM at higher $\alpha$. In our empirical evaluations, we show that these benefits hold across a wide range of tasks including math, code and natural language tasks, and across a diverse set of base LLMs. We will see later that lower abstention rates are due to \methodname' ability to \textit{selectively} abstain on specific inputs by specifying a stopping rule that exceeds our fixed sampling budget.

\noindent {\bf Contributions.} %
The key contribution is the development and  evaluation of the generative prediction sets framework. %
\begin{itemize}
    \item Develop a provable CP method to produce valid prediction sets %
    from a given deep generative model.
    \item \methodname\ is the first CP method for deep generative models that prescribes a stopping rule for sampling without requiring access to samples from the model at test time. %
    \item Empirical evaluation on multiple benchmark datasets with diverse LLMs over text and code. %
    
\end{itemize}

\section{Background and Problem Setup}

This section provides background and describes the formal problem setup.
We use capital letters $(X, Y)$ for random variables and their lowercase $(x, y)$ for specific values. %

\subsection{Conformal Prediction Background}
\vspace{-1ex}

\begin{figure*}[htbp]
    \centering
    \begin{subfigure}{\textwidth}
        \centering
            \resizebox{\textwidth}{!}{%
                \begin{tikzpicture}[
    arrow/.style={->, >=latex},
    font=\small
]
    \node[rotate=90] at (-8.5,0) {\Large {\bf Calibration}};

    \draw[dashed, fill=gray!10] (-8,-2.6) rectangle (1,2.8);
    \draw[dashed, fill=gray!10] (2,-2.6) rectangle (10,2.8);
    
    \node (step1) at (-3.5,0) {\input{gps-overview-figure/cal/step1}};
    \node (step2) at (6,0) {\input{gps-overview-figure/cal/step2}};
    
    \node[below=0.1cm of step1,  align=center] {\large 1. Collect augmented calibration samples};
    \node[below=0.1cm of step2,  align=center] {\large 2. Calibrate $\hat{f}$ on $(X_i, K_i)$ pairs using CP};
    \draw[->, >=latex, line width=4pt] (1,0) -- (2.8,0);

\end{tikzpicture}
            }
    \end{subfigure}
    
    \vspace{0cm}
    
    \begin{subfigure}{\textwidth}
        \centering
            \resizebox{\textwidth}{!}{%
                \begin{tikzpicture}[
    arrow/.style={->, >=latex},
    font=\small
]
    \node[rotate=90] at (-8.5,0) {\Large {\bf Inference}};

    \draw[dashed, fill=blue!5] (-8,-1.7) rectangle (10,1);
    
    \node[draw, fill=orange!20, text width=7cm, align=left, rounded corners] (problem) at (-4,0) {
        After scoring 14 points, Erin now has three times more points than Sara, who scored 8. How many points did Erin have before?
    };
    \node[below=0.25cm of problem.south] {\large Test input $(X_{n+1})$};

    \node[draw, minimum width=3cm, minimum height=1.5cm,fill=white] (compute) at (2,0) {
        $\hat{K}(X_{n+1})=3$
    };
    \node[below=0.14cm of compute.south] {
        \large Compute $\hat{K}(X_{n+1})$
    };

    \node[draw, minimum width=4cm, minimum height=1.5cm,fill=white] (samples) at (7,0) {
        \input{gps-overview-figure/test/sample-set.tex}
    };
    \node[below=0.1cm of samples.south] {
        \large Collect samples from $\colorbox{black}{\textcolor{white}{$\hat{\pi}(y|x)$}}$
    };
    
    \draw[arrow] (problem) -- (compute);
    \draw[arrow] (compute) -- (samples);
    
\end{tikzpicture} 
            }
    \end{subfigure}
    \caption{High-level illustration of the GPS framework:  Calibration approach (top) and Inference procedure (bottom).}
    \label{fig:gps-overview}
\end{figure*}

CP is a general %
framework for constructing prediction sets with valid coverage guarantees \cite{Vovk2005-cp,lei2014distribution}. Split CP is the most widely used variant of CP which works as follows. Given a calibration dataset $\{(X_i, Y_i)\}_{i=1}^n$ (e.g., image and class label pairs), a predictive model $\hat{f}: \mathcal{X} \rightarrow \mathcal{Y}$ (e.g., neural network classifier), a significance level $\alpha \in (0, 1)$, and a new test input $(X_{n+1}, Y_{n+1})$ such that $\{(X_i, Y_i)\}_{i=1}^{n+1}$ are {\em exchangeable}, CP constructs a prediction set $\hat{C}(X_{n+1}) \subset \mathcal{Y}$, such that the set contains $Y_{n+1}$ with $1-\alpha$ probability. 
A key component of CP is the \textit{non-conformity score}, $\mathcal{S}: \mathcal{X} \times \mathcal{Y} \rightarrow \mathbb{R}$, a heuristic measure of the degree to which a classifier's prediction conforms to a given input.
For classification tasks, the output distribution of a classifier can be used as $\mathcal{S}(x, y)$ = $1 - \hat{f}(y| x), y=1,\cdots,|\mathcal{Y}|$.
Let $S_i = \mathcal{S}(X_i, Y_i)$ denote the non-conformity score of the $i$-th calibration example and $\tau$ = $Q_{1-\alpha}(\{S_i\}_{i=1}^n)$ is the conformal $\alpha$-quantile (i.e., the $\ceil{(1-\alpha)(n+1)/n}$ quantile) of the empirical distribution of the scores.
CP defines the prediction set as:
\begin{equation}
\label{eq:conformal-set-classification}
\hat{C}(X_{n+1}) = \{y \in \mathcal{Y}: \mathcal{S}(X_{n+1}, y) \leq \tau\}    
\end{equation}

Let $P$ denote the joint distribution of all $n+1$ samples. CP then provides the following coverage guarantee~\cite{lei2014distribution}:
\begin{equation}
    \label{eq:conformal-guarantee}
    P\{Y_{n+1} \in \hat{C}(X_{n+1})\} \geq 1-\alpha
\end{equation}

If scores are almost surely distinct (e.g., scores are continuous), %
we can obtain an upper bound~\cite{romano2019conformalized}:
\begin{equation}
    \label{eq:conformal-guarantee-ub}
    P\{Y_{n+1} \in \hat{C}(X_{n+1})\} \leq 1-\alpha + \frac{1}{n+1}
\end{equation}
The guarantees above are \textit{distribution-free}, i.e., they hold for any data distribution $P$ for any $\alpha \in (0, 1)$. They are also \textit{marginal}, in the sense that the probability statements above are over the randomness of all $n+1$ inputs (both the calibration data and the test input), and not conditional on either the calibration data or the test input $X_{n+1}$.

\subsection{CP for Open-Ended Generative Tasks}

The non-conformity score $\mathcal{S}(x, y)$ measures the distance between a base predictor's outputs $\hat{f}(x)$ and true labels $y$. For a calibration example $(x, y)$, this score represents the minimum ``radius'' around $\hat{f}(x)$ that encompasses the true label $y$. This interpretation is straightforward in regression for the typical score $\mathcal{S}(x, y) = |y - \hat{f}(x)|$. At test time, computing the appropriate quantile $\tau$ of these distances allows us to construct sets containing all $y$ within $\tau$-distance of $\hat{f}(x)$.
This approach relies critically on our ability to enumerate all possible outputs within a given radius of $\hat{f}(x)$. For tasks with bounded or ordered output spaces, such as multiple-choice questions using LLMs, this enumeration is feasible and the CP procedure can be directly applied~(\cite{kumar2023conformal}). However, open-ended generative tasks involve combinatorial output spaces that have neither of these properties. In such cases, we can only access a restricted subset of the output space $\mathcal{Y}' \subset \mathcal{Y}$ (such as $M$ samples from a given generative model), which introduces two key challenges:

\noindent{\bf Sample coverage:} We cannot guarantee that other valid outputs $y \in \mathcal{Y}$ don't exist outside our restricted sample space $\mathcal{Y}'$. The vanilla CP process fails to account for the sampling procedure during calibration, as it computes scores $\{S_i\}$ without considering whether true labels appear in the accessible subspace of $\mathcal{Y}$ or not. To address this challenge, we can modify our score to explicitly depend on the samples $(y_1', \dots, y_\tau')$ for some stopping rule $\tau$. One example is:%
    \begin{equation}
        \tilde{\mathcal{S}}(x, (y_1', \dots, y_\tau'), y) = \begin{cases}
            \mathcal{S}(x, y), \text{ if } y \in (y_1', ..., y_\tau') \\
            \infty
        \end{cases} 
    \end{equation}
    
    In practice, we typically require $\tau$ to be bounded. Importantly, for any generative model under finite sampling, there exists a minimum achievable error rate $\alpha^*$. For any target $\alpha < \alpha^*$, the only way to maintain valid coverage is to output the entire space $\mathcal{Y}$ on some inputs -- effectively abstaining on this part of the data distribution.

\noindent{\bf Semantic equivalence:} Generative model outputs often have multiple semantically equivalent ``correct'' solutions. For example, different valid sorting algorithms could solve the same programming task. 
This can lead to situations where our ground truth label, $y$, might not be accessible in $\mathcal{Y}'$, but there might still be a $y' \in\mathcal{Y}'$ that is semantically equivalent to $y$.
To handle this challenge, \cite{Quach2023-mq} introduce a binary admissibility function $\mathcal{A}(x, y)$ that evaluates correctness independent of specific reference labels which we adopt in our work also.

\subsection{Problem Setup and Closest Work}

\paragraph{Problem setup.} We are now ready to formally describe our problem setting. We are given a conditional generative model $\hat{\pi}(\cdot|X)$ over the space $\mathcal{X} \times \mathcal{Y}$, and calibration data $\mathcal{D}_{X}$=$\{(X_i)\}_{i=1}^n$ drawn independently from an unknown distribution $P_X$. We assume that the output space $\mathcal{Y}$ is non-enumerable and unordered.
Let $\mathcal{A}: \mathcal{X} \times \mathcal{Y} \rightarrow \{0, 1\}$ denote an admission function that measures the admissibility of a solution $Y \sim \hat{\pi}(Y|X)$. 
For example, in a code generation task, $\mathcal{A}$ could be a function that checks if a generated program passes all test cases. 
Given a test input $X_{n+1}$, our goal is to generate a prediction set $\hat{C}(X_{n+1}) \subset \mathcal{Y}$ %
such that the set contains at least one admissible output $Y$ with high probability. Formally, for a specified significance level $\alpha \in (0, 1)$, we want to provide the following guarantee:
\begin{equation}
\label{eq:gaps-target-guarantee}
P\{ \exists Y \in \hat{C}(X_{n+1}): \mathcal{A}(X_{n+1}, Y) = 1 \} \geq 1-\alpha
\end{equation}
Our goal is to achieve this guarantee in a finite-sample, distribution-free setting, while minimizing the cost (e.g., number of samples or API usage cost) to generate $\hat{C}(X_{n+1})$. %

\noindent {\bf Conformal language modeling.} CLM \cite{Quach2023-mq} is the closest prior work to ours. It constructs prediction sets with a different type of guarantee than vanilla CP. Specifically, given parameters $\alpha, \delta \in (0, 1)$ and test input $X_{n+1}$, CLM constructs a prediction set $\hat{C}(X_{n+1})$ that satisfies:
\begin{equation}
    \label{equation:clm-pac-style}
    \begin{split}
        P(P(\exists Y \in \hat{C}(X_{n+1}): \mathcal{A}(X_{n+1}, Y) = 1 | \mathcal{D}_{\text{cal}}) \\
        \geq 1-\alpha) \geq 1-\delta
    \end{split}
\end{equation}
Here, the inner probability is over draws of $X_{n+1}$ and the outer probability is over draws of the calibration dataset $\mathcal{D}_{\text{cal}}$. This nested probability structure makes direct comparisons between CLM and CP-based methods challenging, as we discuss in Section~\ref{app:equate-clm-gps}.
CLM consists of three key components: a set confidence estimator $\mathcal{F}$, a sample quality estimator $\mathcal{Q}$, and a sample similarity function $\mathcal{S}$, each parameterized by a threshold $\lambda$. The algorithm iteratively builds prediction sets by generating samples from the generative model and adding them to the set if they meet both quality and diversity thresholds (using $\mathcal{Q}$ and $\mathcal{S}$ respectively). This process continues until the set quality threshold is met, as determined by $\mathcal{F}$. The thresholds $(\lambda_1, \lambda_2, \lambda_3)$ that control risk for each component are determined using the Learn-Then-Test framework \cite{angelopoulos2021learn}.

\section{Generative Prediction Sets}
\label{sec:methodology}

This section describes our {\em Generative Prediction Sets} (GPS) framework to construct prediction sets from a given DGM. 

\subsection{Overview and Advantages of GPS}

\paragraph{Overview of \methodname.} \methodname\ constructs adaptive prediction sets with valid coverage guarantees as illustrated in Figure~\ref{fig:gps-overview} (see Algorithm~\ref{alg:gps} for a complete pseudocode). The core idea behind \methodname\ is to reformulate the prediction set construction problem into estimating the number of samples needed to achieve admissible outputs, operating in two key phases.

\begin{algorithm}
\caption{Generative Prediction Sets (GPS) Framework}\label{alg:gps}
\begin{algorithmic}[1]
\Require generative model $\hat{\pi}(\cdot|X)$, calibration data $\mathcal{D}_X$ = $\{X_i\}_{i=1}^n$, admissibility function $\mathcal{A}$, admissibility estimator $\hat{f}$, sampling budget $M$, test input $X_{n+1}$
\Ensure prediction set $\mathcal{C}(X_{n+1})$ 
\vspace{1ex}
\State {\em // Construct augmented calibration data}
\label{alg:gps-construct-data-start}
\State Initialize $\mathcal{D}_{\text{cal}} = \varnothing$
\For{each $X_i \in \mathcal{D}_X$}
    \State $\hat{K}_i \leftarrow$ minimum number of samples to obtain admissible output within the maximum budget $M$
    \State If no admissible output, then $\hat{K}_i \leftarrow M+1$
    \State $\mathcal{D}_{\text{cal}} \leftarrow \mathcal{D}_{\text{cal}} \cup \{(X_i, \hat{K}_i)\}$
\EndFor
\label{alg:gps-construct-data-end}
\vspace{1ex}
\State {\em // Calibration on minimum number of samples} \label{alg:gps-algorithm-step-calibration-start}
\State Execute CP for regression using  $\mathcal{D}_{\text{cal}}$ and estimator $\hat{f}$ 
\vspace{1ex}
\State {\em // Conformal inference on minimum number of samples}
\State Use conformal regressor on $X_{n+1}$ to get $\hat{K}(X_{n+1})$ as per Eq. \ref{eq:method-conformal-set} (i.e., upper-bound of prediction interval)
\State {\em // Generate $\hat{K}(X_{n+1})$ samples to create prediction set}
\State $\mathcal{C}(X_{n+1}) \leftarrow \{ Y_j \sim \hat{\pi}(\cdot | X_{n+1})  \}_{j=1}^{\hat{K}(X_{n+1})}$ 

\State \textbf{return} Prediction set $\mathcal{C}(X_{n+1})$
\end{algorithmic}
\end{algorithm}

First, \methodname\ creates an augmented calibration dataset by sampling multiple outputs from the given generative model for each calibration input $X_i$ in $\mathcal{D}_{X}$ (steps \ref{alg:gps-construct-data-start}-\ref{alg:gps-construct-data-end} in Algorithm~\ref{alg:gps}). For each $X_i$, we sample from $\hat{\pi}(Y|X)$ using an arbitrary, independent sampling procedure to obtain a sequence of samples $\{Y_{ij}\}_{j=1}^M$ up to a maximum budget $M$. We define $K_i$ to be the number of samples needed to obtain an admissible output from $\hat{\pi}$ for input $X_i$. This gives us an augmented calibration dataset, $\{(X_i, K_i)\}_{i=1}^n$. 

Second, \methodname\ develops a conformal procedure around this augmented data (steps \ref{alg:gps-algorithm-step-calibration-start}-\ref{alg:gps-construct-data-end} in Algorithm~\ref{alg:gps}). Notice that if we construct a prediction interval around $K_i$, we will achieve our desired coverage guarantee due to the following equivalence of events:

\begin{equation}
    \label{eq:y-k-space-equivalence}
    \{K_{i} \leq \hat{K}_i\} \Leftrightarrow \left\{ \sum_{j=1}^{\hat{K}_i} \mathcal{A}(X_i, Y_{ij}) > 0 \right\}
\end{equation}

At test time, given a new input $X_{n+1}$, we predict $K_{n+1}$ (using the upper bound of the conformal interval) and sample from $\hat{\pi}(Y|X)$ exactly $K_{n+1}$ times to construct the prediction set $\mathcal{C}(X_{n+1})$, achieving the guarantee stated in Eq. (\ref{eq:gaps-target-guarantee}).

\noindent {\bf Key advantages of GPS.} \methodname\ has two qualitative benefits over CLM. First, \methodname\ reduces the problem of generating prediction sets for DGMs to a vanilla CP problem. This reduction allows us to bring to bear the full body of machinery developed in vanilla CP to this problem setting. For example, the approximate conditional coverage method of \citet{Gibbs2023-ax} can be applied off-the-shelf to \methodname. Achieving such conditional coverage guarantees with CLM would require a non-trivial extension of the LTT framework.  Second, since \methodname\ calibrates only a single parameter, it has substantially lower computational complexity than CLM. Lastly, \methodname\ works in batch mode; it specifies how many samples are to be collected from the DGM apriori (see Figure~\ref{fig:gps-overview}). In contrast, CLM works sequentially, one sample at a time. Thus, \methodname\ can easily be used in modern batch inference pipelines, where it is desirable to produce all samples at once in a single batch to maximize hardware utilization, or in cost-sensitive applications, where users might want to make a trade-off between abstention rate and sampling cost.

\subsection{Conformal Calibration Algorithm}
\label{sec:calibration-algorithm}

A key challenge in designing calibration procedures for sampling-based regimes is handling cases where $\hat{\pi}$ fails to produce an admissible solution within $M$ samples. Prior work in CP for generative models has addressed this by either setting non-conformity scores to $\infty$ (\cite{su2024api}) or assuming such failures don't occur (\cite{wang2024conu}). However, if more than $\ceil{(1-\alpha)(n+1)}$ calibration examples lack admissible solutions, the conformal quantile $Q_{1-\alpha}(\{S_i\}_{i=1}^n)$ becomes $\infty$, yielding trivial prediction sets ($\mathcal{Y}$). \methodname\ addresses this challenge via a different approach.

For calibration examples where no admissible solution is found within the sampling budget, we set $K_i=M+1$, defining $K_i = \inf \{j: \mathcal{A}(X_i, Y_{ij}) = 1\} \cup \{M+1\}$. This sentinel value $M+1$ indicates the absence of an admissible solution in $M$ samples. While the true $K_i$ may be unknown in such cases, we know it exceeds $M$. This definition ensures finite $K_i$ values, enabling the use of any standard regression-based non-conformity scores to calibrate $\{(X_i, K_i)\}_{i=1}^{n}$. %
This choice also allows us to handle abstention in a principled way.

To develop this calibration procedure, we first characterize the distribution of $K_i$. For our problem setting, the samples are collected using an I.I.D. sampling process, and thus, $K_i$ follows a geometric distribution conditional on $X_i$, with its success probability an unknown function of $X_i$, i.e. $K_i$ given $X_i=x$ follows Geom($f(x)$). While $f$ is unknown, we can train an estimator $\hat{f}$ to estimate the success probability conditional on $X_i$, on a separate training split. We then use $\hat{f}$ to produce an estimate of the $1-\alpha$ conditional quantile for $K_{n+1}$, being careful to account for finite sampling:
\begin{equation}\label{eq:qhat}
    \hat{q}_{\alpha}(x) := \ceil{\frac{\ln{\alpha}}{\ln{\{ 1-\hat{f}(x) \}}}}
\end{equation}
While this quantile estimate is simply an estimate and will not yield a $1-\alpha$ coverage in finite samples, we can conformalize it to obtain the valid, finite-sample coverage guarantees. For this, we use the Conformalized Quantile Regression (CQR) score from~\cite{romano2019conformalized}:
$$
S(x, k) = \max\{0, k - \hat{q}_{\alpha}(x)\} 
$$

The score function is asymmetric, since we are building a one-sided interval of the form $[0, \hat{K}]$. Next, using we can construct a one-sided interval for $K_{n+1}$ by applying the conformal adjustment term: %
\begin{equation}
\label{eq:k-estimate}
    \hat{K}(X_{n+1}) = \hat{q}_{\alpha}(X_{n+1}) + Q_{1-\alpha}(\{S_i\}_{i=1}^n) 
\end{equation}

Recall that $\hat{Q}_{1-\alpha}$ is the conformal quantile over calibration scores. 
Now, because of our definition of $K_i$, $\hat{K}$ will always produce finite values, although they might exceed $M$. If this does happen, we must allow the model to abstain; it must output infinite sets. We can do this using the below scheme: %
\begin{equation}
    \label{eq:method-conformal-set}
    \hat{K}(X_{n+1}) = 
    \begin{cases}
        \hat{K}(X_{n+1}) & \text{ , if } \hat{K}(X_{n+1}) \leq M \\
        \infty & \text{ otherwise} \\
    \end{cases}
\end{equation}

With the above $\hat{K}$, we have the below guarantee due to CP:
\begin{equation}
    P\{K_{n+1} \leq \hat{K}(X_{n+1})\} \geq 1-\alpha
\end{equation}
This can then be translated back to the original output space, $\mathcal{Y}$, resulting in the following coverage guarantee for GPS:
\begin{proposition}
Let $\{Y_j \sim \hat{\pi}(\cdot|X=X_{n+1})\}_{j=1}^{\hat{K}(X_{n+1})}$ be the prediction set generated according to Algorithm 1. Then we have the following coverage guarantee:
\begin{equation*}
    P\{\exists Y \in \hat{C}(X_{n+1}): \mathcal{A}(X_{n+1}, Y) = 1\} \geq 1-\alpha
\end{equation*}
\end{proposition}

This statement is a direct result of the equivalence of events stated in Eq.~\ref{eq:y-k-space-equivalence}. We include a proof in the Appendix.

To understand $\hat{K}$'s abstention behavior, consider two extremes. When $\hat{f}$ accurately predicts $\hat{\pi}$'s abstentions, many calibration scores will be small, resulting in a modest conformal adjustment in Eq.~\ref{eq:k-estimate}. Thus, abstention decisions are primarily driven by $\hat{f}$'s input-specific predictions. Conversely, with a poor estimator $\hat{f}$ that outputs a constant probability (yielding $\hat{q}_{\alpha}=1$), calibration examples where $\hat{\pi}$ abstains will have scores of $M$. With enough such examples, this forces $\hat{K}=M+1$ universally, leading to constant abstention. Our calibration procedure thus adapts based on both $\hat{f}$'s quality and $\hat{\pi}$'s abstention rate.

We have now obtained a stopping rule for sampling, $\hat{K}(X_{n+1})$, without having to generate any samples from $\hat{\pi}(\cdot | X)$. At this point, the user has the option to abstain, if $\hat{K}$ is either $\infty$ or deemed too high for their usage scenario. To construct the prediction set from combinatorial space $\mathcal{Y}$, we take $\hat{K}(X_{n+1})$ samples from the generative model:
\begin{equation}
{\hat{C}(X_{n+1})} = \{Y_{j} \sim \hat{\pi}(\cdot|X=X_{n+1})\}_{j=1}^{\hat{K}(X_{n+1})}
\end{equation}
$\hat{C}(X_{n+1})$ contains only distinct outputs. Due to potential duplicate samples from the generative model, the cardinality of $\hat{C}(X_{n+1})$ may be less than or equal to $\hat{K}(X_{n+1})$.

\noindent {\bf Construction of admissibility estimator $\hat{f}$.} 
While \methodname\ guarantees valid prediction sets with the desired coverage level, the admissibility predictor $\hat{f}$ determines three key performance characteristics: empirical set sizes, sampling efficiency, and adaptivity of the prediction sets.

The construction of $\hat{f}$ requires collecting multiple samples from the underlying deep generative model using a separate training set. For each training example $X_i$, we obtain samples $\{Y_{ij}\}_{j=1}^{M_{\text{train}}}$, where $M_{\text{train}}$ need not equal the calibration parameter $M$. The probability of success ($f(X_i)$) can be estimated, for example, using a MAP estimate:
\begin{equation}
    p_i = \frac{\sum_{j=1}^{M_{\text{train}}}\mathcal{A}(X_i, Y_{ij}) + 1}{M_{\text{train}} + 2},
\end{equation}
yielding training data pairs $(X_i, p_i)$.

Training $\hat{f}$ to predict success probabilities requires input signals derived solely from $X_i$. The primary options are log probabilities and latent space representations from the generative model. When log probabilities are inaccessible (as in API-based models), one can utilize either a separate embedding model or hidden states from a smaller surrogate model. In Section~\ref{sec:experimental-results}, we demonstrate this approach using both input log probabilities and hidden states from a smaller model (Phi 2) to train $\hat{f}$ for a larger models (GPT4o-mini). A constant predictor serves as a final alternative when no input signals are available---while preserving marginal validity, this yields fixed-size prediction sets.

\vspace{-2ex}

\section{Experimental Results}
\label{sec:experimental-results}

In this section, we describe the experimental evaluation of our proposed GPS framework, comparison with state-of-the-art methods, and discuss results along different dimensions. 

\vspace{-2ex}
\subsection{Experimental Setup}

\begin{figure*}[!h]
     \centering
        \includegraphics[width=\textwidth]{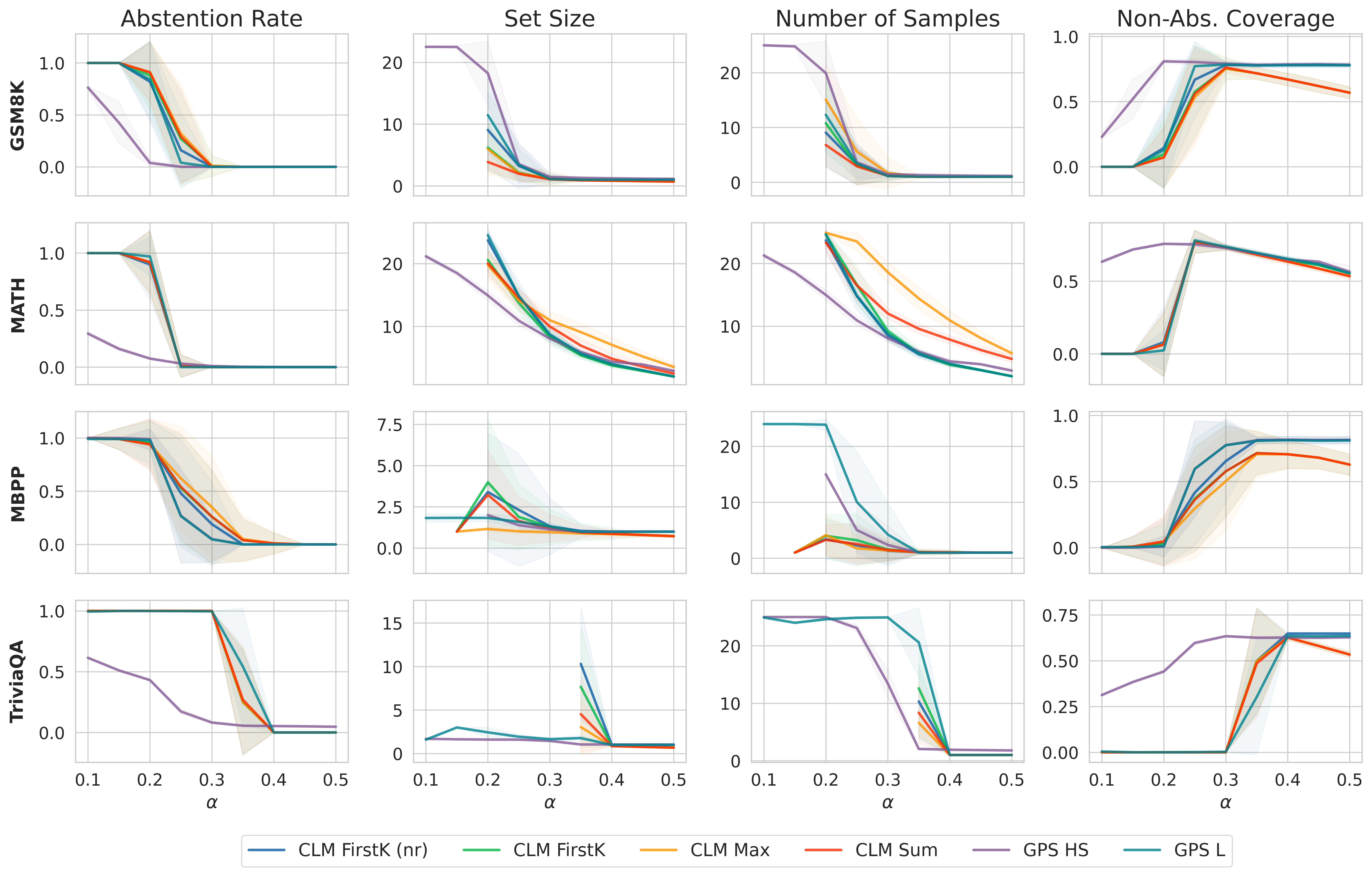}
     \caption{Results comparing GPS and CLM on different datasets (one row each for GSM8K, MATH, MBPP, and TriviaQA) using GPT-4o-mini LLM, across 100 trials, with shaded regions indicating standard deviation. Both \texttt{GPS} variants achieve lower abstention rates, and higher non-abstention coverage, while maintaining competitive set sizes with respect to CLM.} %
     \label{fig:main-results}
\end{figure*}

\noindent {\bf Benchmark datasets.}
We employ data from three diverse tasks. %
{\em 1) Code generation tasks:} we employ the MBPP\cite{Austin2021-qs} and DS-1000~\cite{lai2023ds} datasets to evaluate performance on code generation tasks. We use execution based functional correctness as our binary admission function. {\em 2) Math problem-solving tasks:} We employ the GSM-8K~\cite{Cobbe2021-lg} and Math~\cite{hendrycks2021measuring} datasets on math problem-solving tasks. {\em 3) Natural language understanding tasks:} We employ a natural language question-answering dataset, Trivia-QA~\cite{joshi2017triviaqa}.  For both math problems and natural language understanding, we use exact match as the admissibility function. Since our baseline CLM wasn't able to provide any valid configurations DS-1000 for all $\alpha \in [0.1, 0.5]$, we defer results for DS-1000 to the Appendix. %

\noindent {\bf Deep generative models.} We consider three models across various parameter scales: Phi-2~\cite{li2023textbooks}, Llama 3 8B, and GPT-4o mini. For Math and DS-1000, we also add Gemma 2 27B~\cite{team2024gemma}, since Phi 2 and Llama 3 8B had low success rates for both of these tasks. Across all models, we use nucleus sampling~\cite{holtzman2019curious}, with a sampling budget of 25.
For space reasons, we only present results on GPT-4o mini in the main text, while the results for the other models are provided in the Appendix.

\noindent {\bf Configuration of GPS.} We consider two predictors for the admissibility estimator $\hat{f}$. First, as a baseline, \texttt{GPS-L} uses a a linear regressor that directly predicts the probability of succes using the log probability of the input prompt. On the other hand,\texttt{GPS-HS} uses a feed-forward neural network that takes latent space representation of the prompt as input to predict the probability of success. Since GPT4o mini's hidden states aren't accessible, we use hidden state activations from Phi-2 as a surrogate. In the Appendix, we provide results for models with accessible hidden states.

\noindent {\bf CLM baseline.} We employ CLM by adapting its code\footnote{https://github.com/Varal7/conformal-language-modeling} as our baseline for our experiments, using normalized log probabilities as the quality score ($\mathcal{Q}$), and ROUGE-L as the similarity score ($\mathcal{S}$), and $\delta = 0.05$. For the set quality score, $\mathcal{F}$, we consider the four variants reported in the original paper by \citet{Quach2023-mq}. These variants differ only in their choice of $\mathcal{F}$; \texttt{CLM\ First-K} uses the size of the prediction set, while \texttt{CLM\ Sum} and \texttt{CLM\ Max} use the sum and maximum of the quality scores of each sample in the set as the score. \texttt{CLM\ First-K} (nr) is the same as \texttt{CLM\ First-K}, but without any rejection rule (no rejection).

\noindent {\bf Evaluation methodology.} We evaluate \methodname\ and CLM across $\alpha$ ranging from 0.1 to 0.5 across 100 trials in increments of 0.05. This covers the spectrum of $\alpha$ values that are useful in practice. CLM's guarantees are qualitatively different from that of vanilla CP, so for a fixed $\alpha$, the sets generated by CLM and \texttt{GPS} cannot be directly compared. However, we can use the fact that calibration conditional coverage of conformal predictors follows Beta distribution depending only on $\alpha$ and $n$, to find $\alpha_\delta$, the confidence level at which CP will achieve the same guarantee as CLM. Thus, we follow \citet{Quach2023-mq} and set $\delta=0.05$, and adjust $\alpha$ for GPS to ensure a fair comparison. %

\noindent {\bf Metrics.} We consider four metrics to measure on testing samples: {\bf 1)} average prediction set size, {\bf 2)} average number of samples generated, {\bf 3)} abstention rate, and {\bf 4)} non-abstention empirical coverage. As mentioned previously, the abstention rate allows us to determine the range of $\alpha$ on which a method can produce valid prediction sets. However, for methods such as \methodname\ that can abstain selectively on certain inputs, the abstention rate is not sufficient; for a given $\alpha$ level, a method might simultaneously have a both low abstention and low coverage when it doesn't abstain, i.e. we are not gaining any coverage as a consequence of lowering the abstention rate. Thus, we also present the non-abstention coverage rate: the fraction of data on which a method a) does not abstain, i.e., outputs finite sized sets, and b) contains an admissible solution. Since both CLM and GPS achieve valid empirical coverage, we %
show those results in Appendix. %

\begin{figure*}[!h]
     \centering
     \begin{subfigure}[b]{\columnwidth}
         \centering
         \includegraphics[width=0.9\columnwidth]{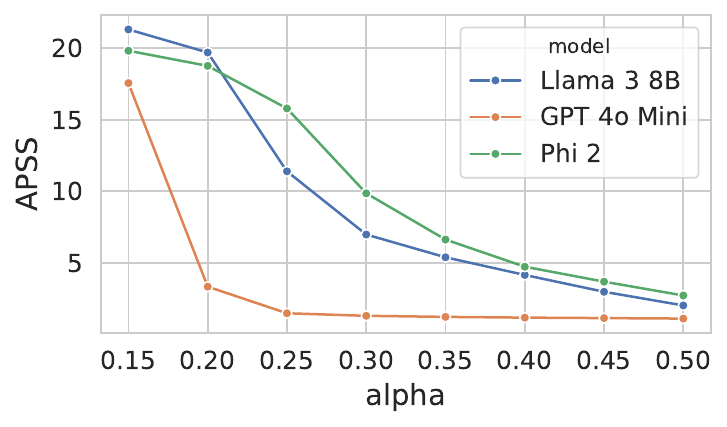}
         \caption{GSM Benchmark}
         \label{fig:scaling-w-pass-rate-gsm}
     \end{subfigure}
     \hfill
     \begin{subfigure}[b]{\columnwidth}
         \centering
         \includegraphics[width=0.9\columnwidth]{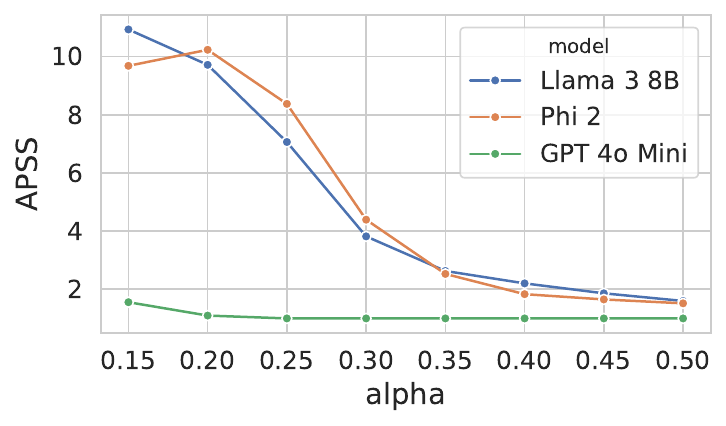}
         \caption{MBPP Benchmark}
         \label{fig:scaling-w-pass-rate-mbpp}
     \end{subfigure}
     \caption{Results for \texttt{GPS-HS}: average prediction set size (APSS) vs. coverage level $\alpha$ for (a) GSM8k and (b) MBPP benchmark datasets, using varying quality of models (Phi-2, Llama 3 8b, and GPT-4o-mini).}
     \vspace{-3mm}
\end{figure*}

\subsection{Results and Discussion}

\noindent {\bf Overall discussion.} To summarize, empirical results in Fig ~\ref{fig:main-results} show that \texttt{GPS} can produce set sizes that are competitive with CLM, but at a wider range of $\alpha$ and with higher non-abstention coverage. We stress that \texttt{GPS} achieves such performance a) \textit{without} ever examining the generated samples, b) using a simpler calibration algorithm that is easy to implement in practice, and c) in $\alpha$-regimes where CLM simply doesn't produce valid configurations. We highlight our key observations from Figure~\ref{fig:main-results} below:
{\bf 1.} \texttt{GPS-L} and \texttt{GPS-HS} achieve higher non-abstention coverage than CLM variants, particularly for $\alpha \leq 0.3$. \texttt{GPS-HS} shows superior non-abstention coverage across all $\alpha$ except in MBPP, likely due to its small training set (150 examples) yielding a poor quality $\hat{f}$. On GSM8k (659 examples), \texttt{GPS HS} consistently outperforms all methods across all $\alpha$ in both abstention rates and non-abstention coverage. \\
{\bf 2.} For GSM, MATH, and TriviaQA, \texttt{GPS HS} maintains abstention rate of $< 1$ and abstention coverage $\geq 0.45$ even at $\alpha \approx 0.1$. This demonstrates $\hat{f}$'s effectiveness at selective abstention on difficult inputs, yielding valid configurations with non-trivial coverage. The predictor quality's impact is clear: \texttt{GPS L} uses same calibration but with a weaker $\hat{f}$. \\
{\bf 3.} \texttt{GPS L} and \texttt{GPS HS} require fewer samples to get prediction sets, especially on MATH. A narrow $\alpha$ range marking the transition between non-zero and zero abstention shows CLM requiring fewer samples than \texttt{GPS}, but with higher abstention rates and lower non-abstention coverage. The sole exception is \texttt{GPS L} on TriviaQA for $\alpha \in [0.35, 0.4]$, where the best CLM method achieves marginally higher non-abstention coverage with fewer samples. \\
{\bf 4.} Both \texttt{GPS} variants produce comparable set sizes to the best CLM method when abstention rates approach 0 ($\alpha \geq 0.3$). At tighter confidence levels, CLM yields smaller sets but with near-zero non-abstention coverage (e.g., GSM8k and MATH at $\alpha=0.2$), while \texttt{GPS L} maintains non-abstention coverage $\geq 0.4$. Set size differences diminish as abstention rates approach zero.

\noindent {\bf Set size vs. model quality.} %
Fig~\ref{fig:scaling-w-pass-rate-gsm} and~\ref{fig:scaling-w-pass-rate-mbpp} show \texttt{GPS-HS} set sizes across model performance levels for MBPP and GSM at different $\alpha$ (see Appendix for more results). Higher model pass rates (problems with at least one admissible solution) correlate with smaller generated sets, demonstrating \methodname's ability to scale with model quality. Fluctuations at higher coverage levels ($\alpha \in \{0.15,0.2\})$ stem from abstention rates, which are higher for Phi-2 than Llama on both datasets, reducing prediction set sizes at these levels.

\section{Related Work}

\noindent {\bf Conformal prediction.} In recent years, CP has emerged as an effective framework for uncertainty quantification across both regression~\cite{romano2019conformalized,papadopoulos2002inductive} and classification tasks~\cite{romano2020classification,angelopoulos2020uncertainty,sadinle2019least}. Much of the recent focus in the CP community has been on reducing set sizes~\cite{romano2020classification,angelopoulos-sets,sadinle2019least} and constructing CP procedures for specific usage contexts~\cite{huang2024uncertainty}. Some works also focus on providing conditional coverage guarantees~\cite{Gibbs2023-ax,jung2022batch,tibshirani2019conformal}. Conformal risk control extends CP to control generalized risk on prediction sets~\cite{angelopoulos2022conformal,bates2021distribution}. The Learn-Then-Test \citet{angelopoulos2021learn} is a framework for risk controlled prediction sets that does not require monotonic risks. It is utilized by CLM~\citet{Quach2023-mq} for generating prediction sets from LLMs.

\noindent {\bf Uncertainty estimation for LLMs.} Uncertainty for LLMs has largely focused on calibrating output probabilities~\cite{jiang2021can,desai2020calibration,lin2023generating,kuhn2023semantic}, verbalization~\cite{kadavath2022language}, and Bayesian approaches~\cite{malinin2020uncertainty, ryabinin2021scaling}. Deep classifiers can exhibit high levels of calibration error~\cite{jiang2021can}. CP is recently utilized for generative models, especially LLMs, in settings with bounded output spaces, such as multiple choice question answering~\cite{kumar2023conformal,zhang2021less,rouzrokh2024conflare,li2024traq}. Lastly, another line of work focuses on selecting a subset of the generated output of an LLM with factuality guarantees~\cite{mohri2024language,cherian2024large}.

\section{Summary}

This paper introduced \methodname, a novel conformal prediction algorithm to produce marginally valid prediction sets for deep generative models with unbounded output spaces. %
Our score formulation allows \methodname\ to selectively abstain on specific examples, allowing it to achieve lower abstention rates and higher non-abstention coverage at $\alpha$-levels at which prior work always abstains. %
Our experiments show that \methodname\ indeed achieves higher non-abstention coverage and lower abstention rates, while maintaining parity with CLM with regard to set sizes at higher $\alpha$ levels. %

\vspace{1ex}

\noindent {\bf Acknowledgements.} The authors gratefully acknowledge the in part support by USDA-NIFA funded AgAID Institute award 2021-67021-35344. The views expressed are those of the authors and do not reflect the official policy or position of the USDA-NIFA.

\bibliography{refs}

\newpage

\onecolumn

\begin{center}
\vspace*{\fill}
{\LARGE\bfseries Appendix}\\[1cm]  %
\vspace*{\fill}
\end{center}
\appendix
\onecolumn
\section{Proofs}
\label{app:proofs}

\subsection{Proof for Prop 3.1}
Recall that $K_{i} = \inf\{j: \mathcal{A}(X_i, Y_{ij}) = 1\} \cup \{M+1\}$, and $\hat{C}(X_{n+1}) = \{Y_{j} \sim \hat{\pi}(\cdot|X=X_{n+1})\}_{j=1}^{\hat{K}(X_{n+1})}$. From the standard CP proof, we have that:
\[
    P\{K_{n+1} \leq \hat{K}(X_{n+1})\} \geq 1-\alpha
\] Consider the event $\{K_{n+1} \leq \hat{K}(X_{n+1})\}$:
\begin{equation}
    \label{app:eq:coverage-event}
        \{K_{n+1} \leq \hat{K}(X_{n+1}) \} = \{K_{n+1} \leq \hat{K}(X_{n+1}) \} \{ \hat{K}(X_{n+1}) < \infty\} \cup \{K_{n+1} \leq \hat{K}(X_{n+1}) \} \{ \hat{K}(X_{n+1} = \infty\} 
\end{equation}

On $\{\hat{K}(X_{n+1}) = \infty\}$, $\hat{C}(X_{n+1}) = \mathcal{Y}$, and $\{\exists Y \in \hat{C}(X_{n+1}): \mathcal{A}(X_{n+1}, Y) = 1\}$. 

On $\{\hat{K}(X_{n+1}) < \infty\}$:
\[
\begin{aligned}
    \{K_{n+1} \leq \hat{K}(X_{n+1}) \} &= \left\{\inf\{j: \mathcal{A}(X_{n+1}, Y_{n+1,j}) = 1\} < \hat{K}(X_{n+1}) \right\} \\
    &= \left\{ \exists j < \hat{K}(X_{n+1}):  \mathcal{A}(X_{n+1}, Y_{n+1,j}) = 1\ \right\} \\
    &= \left\{\exists Y \in \hat{C}(X_{n+1}): \mathcal{A}(X_{n+1}, Y) = 1 \right\} \\
    \end{aligned}
\]
Substituting in Eq.\ref{app:eq:coverage-event}, we obtain:
\[
\{K_{n+1} \leq \hat{K}(X_{n+1})\} = \{\exists Y \in \hat{C}(X_{n+1}): \mathcal{A}(X_{n+1}, Y) = 1\}
\] and plugging this into the probability statement gives us the desired coverage guarantee.

\section{Equating theoretical guarantees between CLM and GPS}
\label{app:equate-clm-gps}
The marginal guarantee of conformal prediction Eq. \ref{eq:conformal-guarantee} and CLM's guarantee Eq. \ref{equation:clm-pac-style}
are qualitatively different. In this paper, whenever we mention $\alpha$ for comparing our method, it refers to $\alpha$ in
Eq. \ref{equation:clm-pac-style}, and the conformal value is adjusted accordingly. We use the calibration conditional coverage properties of CP to perform this adjustment.

Assuming non-conformity scores are almost surely distinct, the calibration coverage, conditional on the calibration set of a conformal predictor at level $\alpha_0$ follows a Beta distribution~(\cite{angelopoulos2024theoretical}:
\begin{equation}
\label{app:eq:calibration-cond-cov}
P\bigl\{Y_{n+1} \in C(X_{n+1}) \mid \mathcal{D}_\text{cal} \bigr\} \sim \text{Beta}(k(\alpha_0), n + 1 - k(\alpha_0))
\end{equation}
where $k(\alpha_0) := \left \lceil (1 - \alpha_0) (n + 1) \right \rceil$. Even if scores are not distinct, the miscoverage probability stochastically dominates the Beta$(k_\alpha, n+1-k_\alpha)$ distribution~(\cite{angelopoulos2024theoretical}[Theorem 4.1]).

Following that, for any value of $\alpha$, this conformal method has the following guarantee:
\[
P\Bigl(P(\exists Y \in \hat{C}(X_{n+1}): \mathcal{A}(X_{n+1}, Y) = 1 \mid \mathcal{D}_{\text{cal}})
\geq 1-\alpha\Bigr) \geq 1-\delta
\]

where $\delta$ is the CDF of the Beta distribution with parameters from the result stated in Eq\ref{app:eq:calibration-cond-cov} at $1 - \alpha$:
\begin{equation}
\label{appendix:equation:beta-cdf}
\delta = \text{BetaCDF}_{k(\alpha_0), n + 1 - k(\alpha_0)}(1 - \alpha)
\end{equation}

Thus, for any $\alpha$ and $\delta$ satisfying the LTT guarantee with the form of Eq. \ref{equation:clm-pac-style}, we perform a grid search in the $\alpha_0$ space to find values such that the $\delta$ in Eq. \ref{appendix:equation:beta-cdf} is lower than the $\delta$ in Eq. \ref{equation:clm-pac-style}. Table \ref{appendix:table:alpha_values} shows the adjusted confidence levels,  $\alpha_0$, corresponding to the non-adjusted $\alpha$-values in Figure \ref{fig:main-results}.

\begin{figure}[!h]
    \centering
    \begin{center}
    \begin{tabular}{|c|c|c|c|c|c|}
        \hline
        \multirow{1}{*}{$\alpha$} & \multicolumn{1}{c|}{DS1000 ($n=200$)} & \multicolumn{1}{c|}{GSM8k ($n=330$)} & \multicolumn{1}{c|}{MATH ($n=1250$)} & \multicolumn{1}{c|}{MBPP ($n=113$)} & \multicolumn{1}{c|}{TriviaQA ($n=4486$)} \\
        \hline
        0.10 & 0.0594  & 0.0662  & 0.0814  & 0.0526  & 0.0900  \\
        0.15 & 0.0990  & 0.1084  & 0.1277  & 0.0877  & 0.1381  \\
        0.20 & 0.1485  & 0.1566  & 0.1757  & 0.1228  & 0.1867  \\
        0.25 & 0.1881  & 0.1987  & 0.2236  & 0.1754  & 0.2352  \\
        0.30 & 0.2376  & 0.2469  & 0.2715  & 0.2105  & 0.2843  \\
        0.35 & 0.2772  & 0.2951  & 0.3194  & 0.2631  & 0.3337  \\
        0.40 & 0.3267  & 0.3433  & 0.3690  & 0.2982  & 0.3832  \\
        0.45 & 0.3762  & 0.3915  & 0.4185  & 0.3508  & 0.4331  \\
        0.50 & 0.4257  & 0.4397  & 0.4680  & 0.4035  & 0.4830  \\
        \hline
    \end{tabular}
    \end{center}
    \caption{Table of $\alpha$ values and corresponding adjusted $\alpha_0$ for the different tasks used in this paper.}    \label{appendix:table:alpha_values}
\end{figure}

\section{Experimental Details}
\label{app:experiment}

\subsection{Compute Infrastructure and Software}

All inference except for GPT-4o-mini is performed on a single node with 3 H100 GPUs (80GB VRAM), 8 CPUs and 400GB RAM. We utilize \texttt{vllm}\footnote{https://github.com/vllm-project/vllm} to generate samples, and perform a separate forward pass using Huggingface Transformers\footnote{https://huggingface.co/docs/transformers/en/index} to collect logits and hidden states. GPT-4o-mini inference is conducted using the OpenAI client library\footnote{https://github.com/openai/openai-python} for python. 

\subsection{Dataset Details}
All the datasets we use are publicly available from the Hugginface Hub. For GSM\footnote{https://huggingface.co/datasets/openai/gsm8k} we use the test split from \texttt{main} subset. For Math\footnote{https://huggingface.co/datasets/EleutherAI/hendrycks\_math} we use the test split. For TriviaQA\footnote{https://huggingface.co/datasets/mandarjoshi/trivia\_qa} we use the validation split in the \texttt{rca.nocontext} subset of the data. DS1000\footnote{https://huggingface.co/datasets/xlangai/DS-1000} has only a single split, and we use the entire dataset. For MBPP\footnote{https://huggingface.co/datasets/google-research-datasets/mbpp}, we use both the train and test splits from the \texttt{sanitized} subset, due to the small number of examples in the data. Details regarding the size of each split and generation settings are shown in Table~\ref{fig:data-details}.    
\begin{figure}[!h]
    \centering
    \begin{center}
\begin{tabular}{ |c||c|c|c|c| } 
\hline
\textbf{Dataset} & \textbf{Train Split Size} & \textbf{Test Split Size} & \textbf{Temperature} & \textbf{Top-p} \\
 \hline
DS1000 & 600 & 400 & 0.2 & 0.95 \\
MBPP & 150 & 227 & 0.2 & 0.95 \\
TriviaQA & 8972 & 8972 & 0.2 & 0.95 \\
GSM8k & 659 & 660 & 0.2 & 0.95 \\
Math & 2500 & 2500 & 0.6 & 0.95 \\
\hline
\end{tabular}
\end{center}
    \caption{Dataset splits and generation settings.}
    \label{fig:data-details}
\end{figure}

\subsubsection{Prompts} Here we list the prompts used for our experiments. 
Text enclosed in \textcolor{violet}{\{\dots\}} refers to columns in the original dataset.
\textcolor{teal}{teal} colored text denotes the system prompt used by GPT-4o-mini. \textcolor{red}{red} colored text are only added for Gemma 2 27b.

\lstset{
  basicstyle=\ttfamily\small,
  breaklines=true,
  frame=single,
  rulecolor=\color{black},
  backgroundcolor=\color{gray!10},
  commentstyle=\color{green!50!black},
  keywordstyle=\color{blue},
  stringstyle=\color{red},
   showlines=true,
   escapeinside={(*@}{@*)},
  moredelim=**[is][\color{teal}]{@gpt@}{@end@},  %
  moredelim=**[is][\color{red}]{@gemma@}{@end@},  %
  moredelim=**[is][\color{violet}]{@bold@}{@end@},  %
}

\begin{lstlisting}[caption=Prompt for MBPP]
@gpt@Answer the following question. In your response, only write the raw code, do not use markdown and do not add explanations.@end@

@bold@{description}@end@
@bold@{test_list[0]}@end@

\end{lstlisting}

\begin{lstlisting}[caption=Prompt for DS-1000]
@gpt@Answer the following question. In your response, only write the raw code, do not use markdown and do not add explanations@end@
@bold@{prompt}@end@
\end{lstlisting}

\begin{lstlisting}[caption=Prompt for GSM]
Q: @bold@{question}@end@
A: Let's think step by step.
\end{lstlisting}

\begin{lstlisting}[caption=Prompt for Trivia]
Answer these questions:
Q: Which former major league baseball pitcher, known as The Big Unit, now pitches Geico?
A: Randy Johnson
Q: The Philippines were named after which king of Spain?
A: King Philip II
Q: US Vice-President Joe Biden represents which state?
A: DELAWARE
Q: Which, now defunct, political party was founded by Declan Ganley in April 2009?
A: Libertas Ireland
Q: Sept 30, 1966 saw the public unveiling of which popular model of Boeing aircraft?
A: 747
Q: @bold@{question}@end@
A:
\end{lstlisting}

\begin{lstlisting}[caption=Prompt for Math]
@gemma@Answer the following questions. Your final answer should always follow the same format: 'Final Answer: The final answer is [answer]. I hope it is correct.@end@
@gpt@Answer the following questions. Use the answer format provided in the examples. All of your latex expressions must be wrapped in $...$ (for example, to write 'x=2' as latex, write $x=2$).@end@
Problem:
Find the domain of the expression  $\frac{\sqrt{x-2}}{\sqrt{5-x}}$.}

Solution: The expressions inside each square root must be non-negative. Therefore, $x-2 \ge 0$, so $x\ge2$, and $5 - x \ge 0$, so $x \le 5$. Also, the denominator cannot be equal to zero, so $5-x>0$, which gives $x<5$. Therefore, the domain of the expression is $\boxed{[2,5)}$.
Final Answer: The final answer is $[2,5)$. I hope it is correct.

Problem:
If $\det \mathbf{A} = 2$ and $\det \mathbf{B} = 12,$ then find $\det (\mathbf{A} \mathbf{B}).$

Solution: We have that $\det (\mathbf{A} \mathbf{B}) = (\det \mathbf{A})(\det \mathbf{B}) = (2)(12) = \boxed{24}.$
Final Answer: The final answer is $24$. I hope it is correct.

Problem:
Terrell usually lifts two 20-pound weights 12 times. If he uses two 15-pound weights instead, how many times must Terrell lift them in order to lift the same total weight?

Solution: If Terrell lifts two 20-pound weights 12 times, he lifts a total of $2\cdot 12\cdot20=480$ pounds of weight.  If he lifts two 15-pound weights instead for $n$ times, he will lift a total of $2\cdot15\cdot n=30n$ pounds of weight.  Equating this to 480 pounds, we can solve for $n$:
\begin{align*}
30n&=480\
\Rightarrow\qquad n&=480/30=\boxed{16}
\end{align*}
Final Answer: The final answer is $16$. I hope it is correct.

Problem:
If the system of equations

\begin{align*}
6x-4y&=a,\
6y-9x &=b.
\end{align*}has a solution $(x, y)$ where $x$ and $y$ are both nonzero,
find $\frac{a}{b},$ assuming $b$ is nonzero.

Solution: If we multiply the first equation by $-\frac{3}{2}$, we obtain

$$6y-9x=-\frac{3}{2}a.$$Since we also know that $6y-9x=b$, we have

$$-\frac{3}{2}a=b\Rightarrow\frac{a}{b}=\boxed{-\frac{2}{3}}.$$
Final Answer: The final answer is $-\frac{2}{3}$. I hope it is correct.

Problem: @bold@{problem}@end@
Answer:
\end{lstlisting}

\subsection{Results}

We provide extended evaluation results for \methodname\ and CLM for $\alpha \in [0.05, 0.1, 0.15,.2, .25, .3, .35, .4, .45, .5]$ in the figures below. Additional models include Llama 3 8b, Phi-2, and two variants of Gemma 2 27b - the base model and the instruction tuned version. The figures below follow the same format as Figure~\ref{fig:main-results}.

For \texttt{\methodname-L}, we use a linear regression model from \texttt{scikit-learn}, that takes as input the length normalized log probability of the input prompt under the base LLM. For \texttt{\methodname-HS}, we use a multi-layer perceptron implemented in \texttt{torch} that takes hidden state activations of the last token in the input prompt from the last layer of either the underlying model in the case of Llama and Phi-2. For Gemma 2 and GPT-4o-mini, we use hidden state activations from the last token and layer of Phi-2. We perform a grid-search to determine the number of layers ([1, 2, 4]), number of hidden units ([256, 512, 1024]) and training epochs ([15, 20, 25]). Details are provided in the supplementary code. 

\subsubsection{Figures}
We provide plots similar to Figure~\ref{fig:main-results} for each deep generative model and dataset.
\begin{figure*}[!h]
     \centering
     \includegraphics[width=\textwidth]{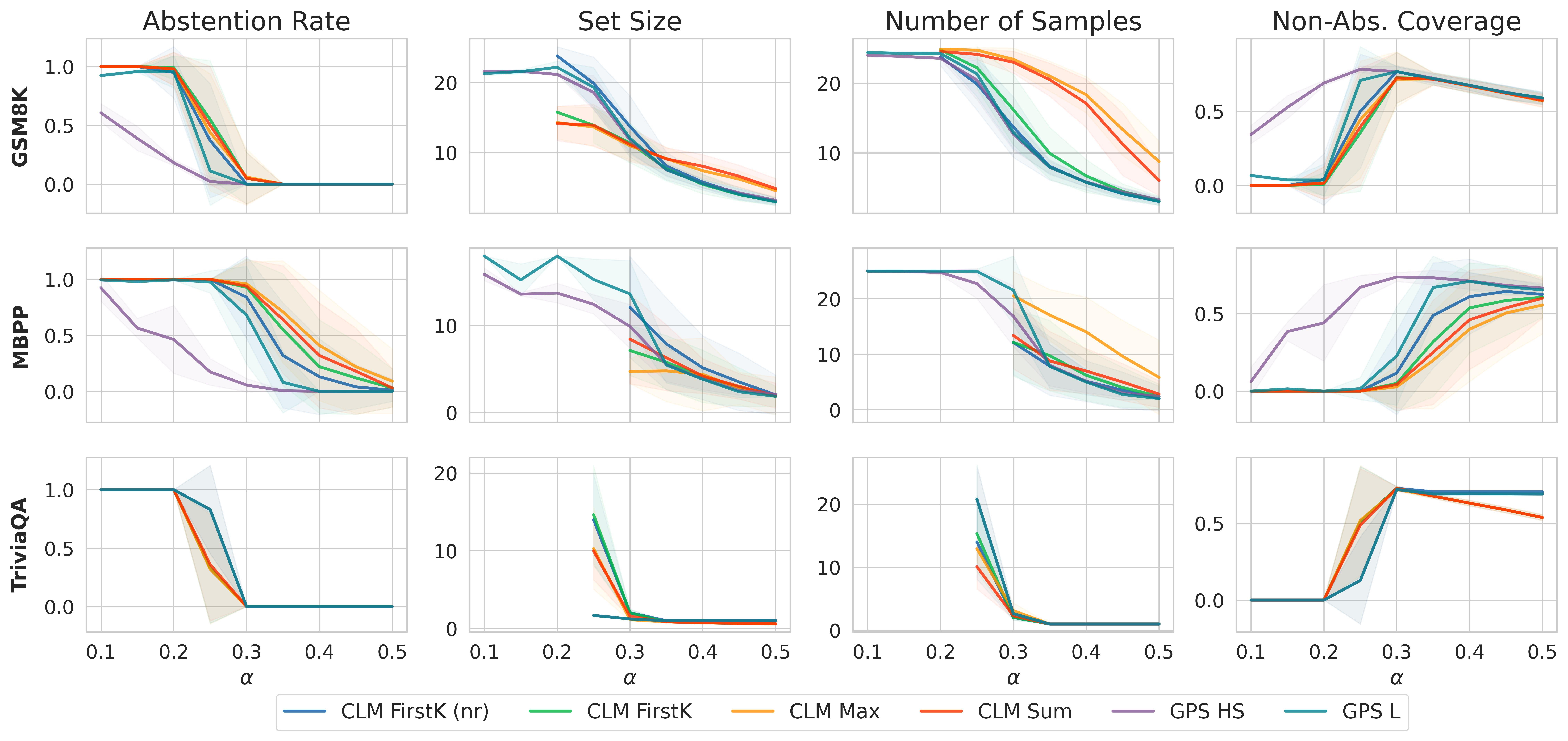}
     \caption{Llama 3 8b results on all datasets.}
\end{figure*}

\begin{figure*}[!h]
     \centering
     \includegraphics[width=\textwidth]{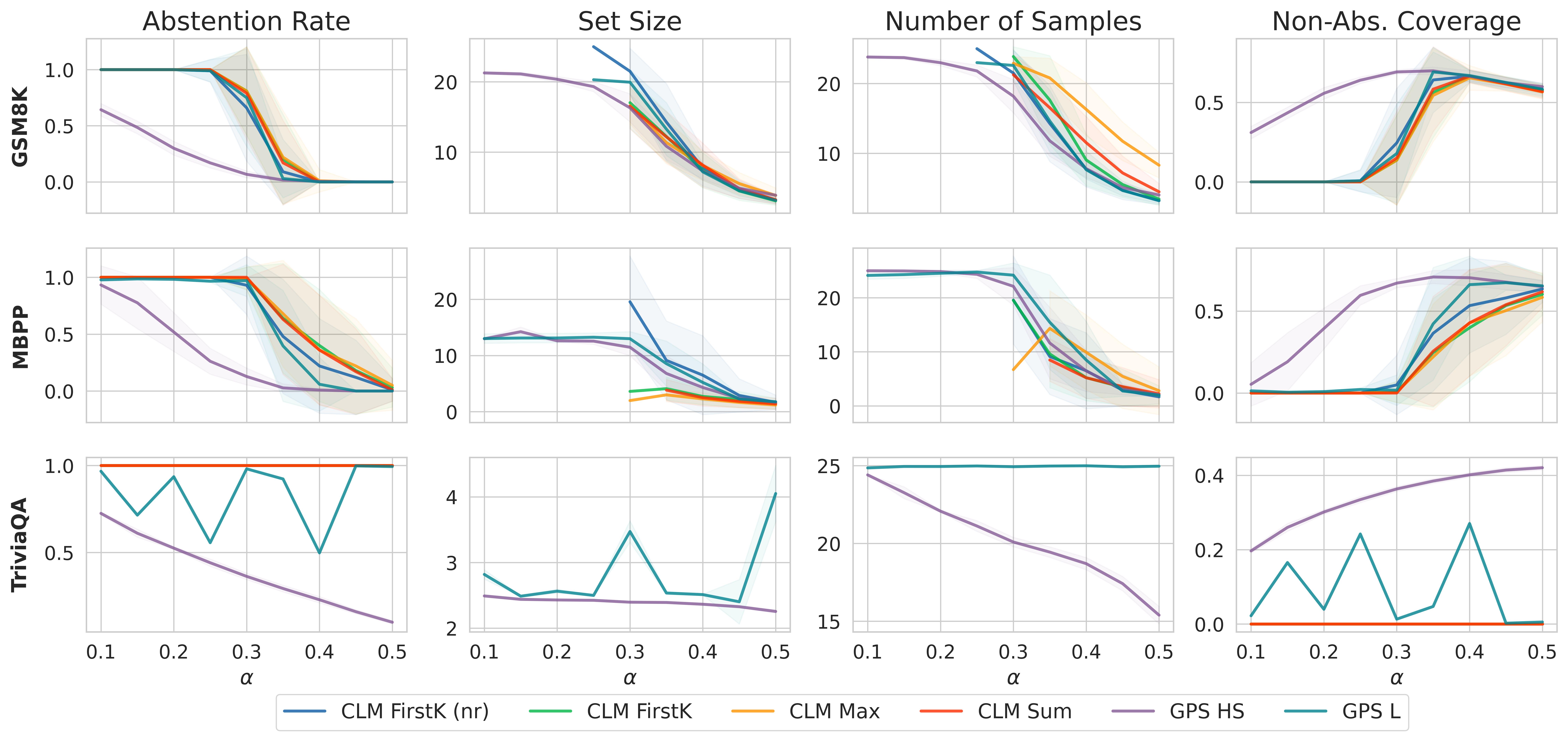}
     \caption{Phi-2 results on all datasets.}
     \vspace{-3mm}
\end{figure*}

\begin{figure*}[!h]
     \centering
     \includegraphics[width=\textwidth]{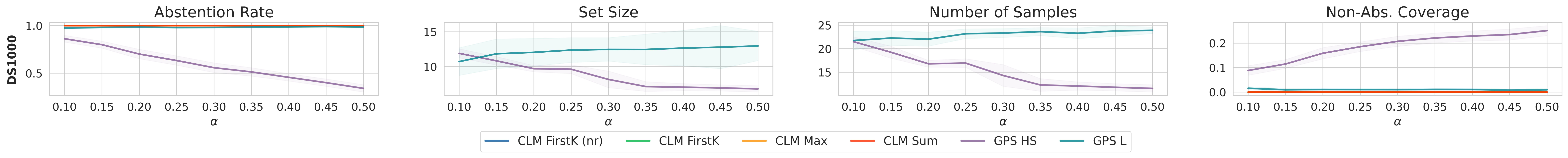}
     \caption{Gemma 2 27b results on DS-1000}
     \vspace{-3mm}
\end{figure*}

\begin{figure*}[!h]
     \centering
     \includegraphics[width=\textwidth]{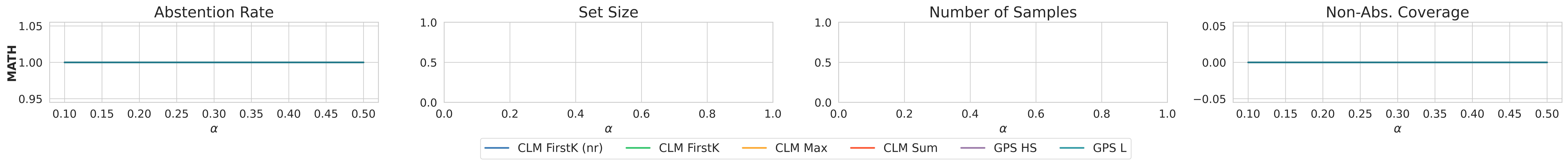}
     \caption{Gemma 2 27b (Instruction Tuned) on DS-1000}
     \vspace{-3mm}
\end{figure*}

\begin{figure*}[!h]
     \centering
     \includegraphics[width=\textwidth]{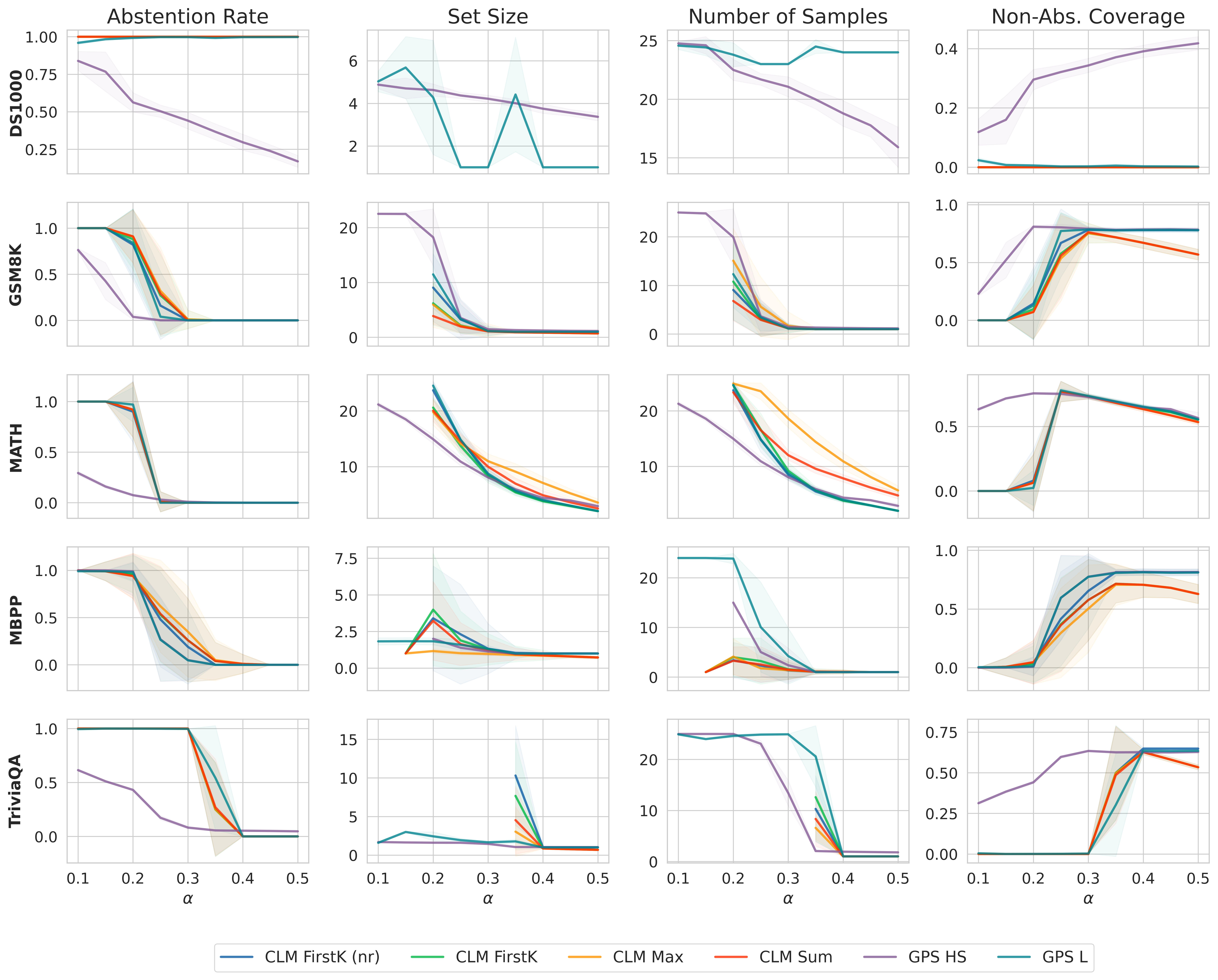}
     \caption{GPT-4o-mini results on all datasets.}
\end{figure*}

\subsubsection{Tables}

Here we provide a comprehensive set of tables including empirical coverage rates, in addition to the metrics considered prior. Blank entries (-) indicate complete abstention.
\begin{figure}[htbp]
    \centering
    \resizebox{\textwidth}{!}{  %
\begin{tabular}{lllllll}
\toprule
 &  & Set Coverage & Abstention Rate & Non-Abs. Coverage & Set Size & Number of Samples \\
$\alpha$ & Method &  &  &  &  &  \\
\midrule
\multirow[t]{6}{*}{0.100000} & CLM FirstK & 1.00±0.00 & 1.00±0.00 & 0.00±0.00 & - & - \\
 & CLM FirstK (nr) & 1.00±0.00 & 1.00±0.00 & 0.00±0.00 & - & - \\
 & CLM Max & 1.00±0.00 & 1.00±0.00 & 0.00±0.00 & - & - \\
 & CLM Sum & 1.00±0.00 & 1.00±0.00 & 0.00±0.00 & - & - \\
 & GPS HS & 0.95±0.02 & \textbf{0.86±0.04} & \textbf{0.17±0.12} & \textbf{21.49±1.04} & 11.90±0.65 \\
 & GPS L & 0.99±0.00 & 0.98±0.01 & 0.12±0.15 & 21.75±1.96 & \textbf{10.72±1.97} \\
\bottomrule
\end{tabular}

    }
    \caption{Results for Gemma 2 27B on DS1000}
\end{figure}
\begin{figure}[htbp]
    \centering
    \resizebox{\textwidth}{!}{  %
        \begin{tabular}{lllllll}
\toprule
 &  & Set Coverage & Abstention Rate & Non-Abs. Coverage & Set Size & Number of Samples \\
$\alpha$ & Method &  &  &  &  &  \\
\midrule
\multirow[t]{6}{*}{0.100000} & CLM FirstK & 1.00±0.00 & 1.00±0.00 & 0.00±0.00 & - & - \\
 & CLM FirstK (nr) & 1.00±0.00 & 1.00±0.00 & 0.00±0.00 & - & - \\
 & CLM Max & 1.00±0.00 & 1.00±0.00 & 0.00±0.00 & - & - \\
 & CLM Sum & 1.00±0.00 & 1.00±0.00 & 0.00±0.00 & - & - \\
 & GPS HS & 0.96±0.02 & \textbf{0.84±0.06} & \textbf{0.23±0.16} & 24.76±0.15 & \textbf{4.88±0.18} \\
 & GPS L & 0.98±0.01 & 0.96±0.01 & 0.17±0.20 & \textbf{24.57±0.33} & 5.03±0.48 \\
\bottomrule
\end{tabular}

    }
    \caption{Results for GPT 4o mini on DS1000}
\end{figure}
\begin{figure}[htbp]
    \centering
    \resizebox{\textwidth}{!}{  %
        \begin{tabular}{lllllll}
\toprule
 &  & Set Coverage & Abstention Rate & Non-Abs. Coverage & Set Size & Number of Samples \\
$\alpha$ & Method &  &  &  &  &  \\
\midrule
\multirow[t]{6}{*}{0.100000} & CLM FirstK & 1.00±0.00 & 1.00±0.00 & 0.00±0.00 & - & - \\
 & CLM FirstK (nr) & 1.00±0.00 & 1.00±0.00 & 0.00±0.00 & - & - \\
 & CLM Max & 1.00±0.00 & 1.00±0.00 & 0.00±0.00 & - & - \\
 & CLM Sum & 1.00±0.00 & 1.00±0.00 & 0.00±0.00 & - & - \\
 & GPS HS & 0.99±0.00 & \textbf{0.76±0.02} & \textbf{0.43±0.28} & \textbf{25.00±0.00} & \textbf{22.51±0.09} \\
 & GPS L & 1.00±0.00 & 1.00±0.00 & 0.28±0.39 & - & - \\
\bottomrule
\end{tabular}

    }
    \caption{Results for GPT 4o mini on GSM8K}
\end{figure}
\begin{figure}[htbp]
    \centering
    \resizebox{\textwidth}{!}{  %
        \begin{tabular}{lllllll}
\toprule
 &  & Set Coverage & Abstention Rate & Non-Abs. Coverage & Set Size & Number of Samples \\
$\alpha$ & Method &  &  &  &  &  \\
\midrule
\multirow[t]{6}{*}{0.100000} & CLM FirstK & 1.00±0.00 & 1.00±0.00 & 0.00±0.00 & - & - \\
 & CLM FirstK (nr) & 1.00±0.00 & 1.00±0.00 & 0.00±0.00 & - & - \\
 & CLM Max & 1.00±0.00 & 1.00±0.00 & 0.00±0.00 & - & - \\
 & CLM Sum & 1.00±0.00 & 1.00±0.00 & 0.00±0.00 & - & - \\
 & GPS HS & 0.95±0.02 & \textbf{0.61±0.08} & \textbf{0.50±0.23} & \textbf{24.02±0.21} & 21.61±0.12 \\
 & GPS L & 0.99±0.00 & 0.92±0.01 & 0.32±0.35 & 24.42±0.16 & \textbf{21.27±0.25} \\
\bottomrule
\end{tabular}

    }
    \caption{Results for Llama 3 8B on GSM8K}
\end{figure}
\begin{figure}[htbp]
    \centering
    \resizebox{\textwidth}{!}{  %
        \begin{tabular}{lllllll}
\toprule
 &  & Set Coverage & Abstention Rate & Non-Abs. Coverage & Set Size & Number of Samples \\
$\alpha$ & Method &  &  &  &  &  \\
\midrule
\multirow[t]{6}{*}{0.100000} & CLM FirstK & 1.00±0.00 & 1.00±0.00 & 0.00±0.00 & - & - \\
 & CLM FirstK (nr) & 1.00±0.00 & 1.00±0.00 & 0.00±0.00 & - & - \\
 & CLM Max & 1.00±0.00 & 1.00±0.00 & 0.00±0.00 & - & - \\
 & CLM Sum & 1.00±0.00 & 1.00±0.00 & 0.00±0.00 & - & - \\
 & GPS HS & 0.95±0.02 & \textbf{0.64±0.06} & \textbf{0.45±0.21} & \textbf{23.81±0.15} & \textbf{21.27±0.15} \\
 & GPS L & 1.00±0.00 & 1.00±0.00 & 0.25±0.35 & - & - \\
\bottomrule
\end{tabular}

    }
    \caption{Results for Phi 2 on GSM8K}
\end{figure}
\begin{figure}[htbp]
    \centering
    \resizebox{\textwidth}{!}{  %
        \begin{tabular}{lllllll}
\toprule
 &  & Set Coverage & Abstention Rate & Non-Abs. Coverage & Set Size & Number of Samples \\
$\alpha$ & Method &  &  &  &  &  \\
\midrule
\multirow[t]{6}{*}{0.100000} & CLM FirstK & 1.00±0.00 & \textbf{1.00±0.00} & 0.00±0.00 & - & - \\
 & CLM FirstK (nr) & 1.00±0.00 & \textbf{1.00±0.00} & 0.00±0.00 & - & - \\
 & CLM Max & 1.00±0.00 & \textbf{1.00±0.00} & 0.00±0.00 & - & - \\
 & CLM Sum & 1.00±0.00 & \textbf{1.00±0.00} & 0.00±0.00 & - & - \\
 & GPS HS & 1.00±0.00 & \textbf{1.00±0.00} & \textbf{0.10±0.14} & - & - \\
 & GPS L & 1.00±0.00 & \textbf{1.00±0.00} & \textbf{0.10±0.14} & - & - \\
\bottomrule
\end{tabular}

    }
    \caption{Results for Gemma 27B-IT on MATH}
\end{figure}
\begin{figure}[htbp]
    \centering
    \resizebox{\textwidth}{!}{  %
        \begin{tabular}{lllllll}
\toprule
 &  & Set Coverage & Abstention Rate & Non-Abs. Coverage & Set Size & Number of Samples \\
$\alpha$ & Method &  &  &  &  &  \\
\midrule
\multirow[t]{6}{*}{0.100000} & CLM FirstK & 1.00±0.00 & 1.00±0.00 & 0.00±0.00 & - & - \\
 & CLM FirstK (nr) & 1.00±0.00 & 1.00±0.00 & 0.00±0.00 & - & - \\
 & CLM Max & 1.00±0.00 & 1.00±0.00 & 0.00±0.00 & - & - \\
 & CLM Sum & 1.00±0.00 & 1.00±0.00 & 0.00±0.00 & - & - \\
 & GPS HS & 0.93±0.01 & \textbf{0.29±0.02} & \textbf{0.69±0.08} & \textbf{21.28±0.39} & \textbf{21.16±0.39} \\
 & GPS L & 1.00±0.00 & 1.00±0.00 & 0.27±0.38 & - & - \\
\bottomrule
\end{tabular}

    }
    \caption{Results for GPT 4o mini on MATH}
\end{figure}
\begin{figure}[htbp]
    \centering
    \resizebox{\textwidth}{!}{  %
        \begin{tabular}{lllllll}
\toprule
 &  & Set Coverage & Abstention Rate & Non-Abs. Coverage & Set Size & Number of Samples \\
$\alpha$ & Method &  &  &  &  &  \\
\midrule
\multirow[t]{6}{*}{0.100000} & CLM FirstK & 1.00±0.00 & 1.00±0.00 & 0.00±0.00 & - & - \\
 & CLM FirstK (nr) & 1.00±0.00 & 1.00±0.00 & 0.00±0.00 & - & - \\
 & CLM Max & 1.00±0.00 & 1.00±0.00 & 0.00±0.00 & - & - \\
 & CLM Sum & 1.00±0.00 & 1.00±0.00 & 0.00±0.00 & - & - \\
 & GPS HS & 1.00±0.00 & 1.00±0.00 & 0.28±0.40 & - & - \\
 & GPS L & 1.00±0.00 & \textbf{0.99±0.01} & \textbf{0.28±0.40} & \textbf{24.00±0.00} & \textbf{1.82±0.24} \\
\bottomrule
\end{tabular}

    }
    \caption{Results for GPT 4o mini on MBPP}
\end{figure}
\begin{figure}[htbp]
    \centering
    \resizebox{\textwidth}{!}{  %
        \begin{tabular}{lllllll}
\toprule
 &  & Set Coverage & Abstention Rate & Non-Abs. Coverage & Set Size & Number of Samples \\
$\alpha$ & Method &  &  &  &  &  \\
\midrule
\multirow[t]{6}{*}{0.100000} & CLM FirstK & 1.00±0.00 & 1.00±0.00 & 0.00±0.00 & - & - \\
 & CLM FirstK (nr) & 1.00±0.00 & 1.00±0.00 & 0.00±0.00 & - & - \\
 & CLM Max & 1.00±0.00 & 1.00±0.00 & 0.00±0.00 & - & - \\
 & CLM Sum & 1.00±0.00 & 1.00±0.00 & 0.00±0.00 & - & - \\
 & GPS HS & 0.99±0.03 & \textbf{0.92±0.12} & \textbf{0.30±0.34} & \textbf{25.00±0.02} & \textbf{15.90±0.72} \\
 & GPS L & 1.00±0.00 & 1.00±0.00 & 0.26±0.37 & 25.00±0.00 & 18.00±0.00 \\
\bottomrule
\end{tabular}

    }
    \caption{Results for Llama 3 8B on MBPP}
\end{figure}
\begin{figure}[htbp]
    \centering
    \resizebox{\textwidth}{!}{  %
        \begin{tabular}{lllllll}
\toprule
 &  & Set Coverage & Abstention Rate & Non-Abs. Coverage & Set Size & Number of Samples \\
$\alpha$ & Method &  &  &  &  &  \\
\midrule
\multirow[t]{6}{*}{0.100000} & CLM FirstK & 1.00±0.00 & 1.00±0.00 & 0.00±0.00 & - & - \\
 & CLM FirstK (nr) & 1.00±0.00 & 1.00±0.00 & 0.00±0.00 & - & - \\
 & CLM Max & 1.00±0.00 & 1.00±0.00 & 0.00±0.00 & - & - \\
 & CLM Sum & 1.00±0.00 & 1.00±0.00 & 0.00±0.00 & - & - \\
 & GPS HS & 0.99±0.04 & \textbf{0.93±0.17} & \textbf{0.28±0.35} & 25.00±0.00 & \textbf{12.99±0.13} \\
 & GPS L & 0.99±0.01 & 0.98±0.01 & 0.26±0.35 & \textbf{24.14±0.68} & 13.02±0.84 \\
\bottomrule
\end{tabular}

    }
    \caption{Results for Phi 2 on MBPP}
\end{figure}
\begin{figure}[htbp]
    \centering
    \resizebox{\textwidth}{!}{  %
        \begin{tabular}{lllllll}
\toprule
 &  & Set Coverage & Abstention Rate & Non-Abs. Coverage & Set Size & Number of Samples \\
$\alpha$ & Method &  &  &  &  &  \\
\midrule
\multirow[t]{6}{*}{0.100000} & CLM FirstK & 1.00±0.00 & 1.00±0.00 & 0.00±0.00 & - & - \\
 & CLM FirstK (nr) & 1.00±0.00 & 1.00±0.00 & 0.00±0.00 & - & - \\
 & CLM Max & 1.00±0.00 & 1.00±0.00 & 0.00±0.00 & - & - \\
 & CLM Sum & 1.00±0.00 & 1.00±0.00 & 0.00±0.00 & - & - \\
 & GPS HS & 0.93±0.00 & \textbf{0.61±0.01} & \textbf{0.43±0.17} & 25.00±0.00 & 1.69±0.01 \\
 & GPS L & 1.00±0.00 & 0.99±0.00 & 0.23±0.31 & \textbf{24.92±0.08} & \textbf{1.60±0.14} \\
\bottomrule
\end{tabular}

    }
    \caption{Results for GPT 4o mini on TriviaQA}
\end{figure}
\begin{figure}[htbp]
    \centering
    \resizebox{\textwidth}{!}{  %
        \begin{tabular}{lllllll}
\toprule
 &  & Set Coverage & Abstention Rate & Non-Abs. Coverage & Set Size & Number of Samples \\
$\alpha$ & Method &  &  &  &  &  \\
\midrule
\multirow[t]{6}{*}{0.100000} & CLM FirstK & 1.00±0.00 & \textbf{1.00±0.00} & 0.00±0.00 & - & - \\
 & CLM FirstK (nr) & 1.00±0.00 & \textbf{1.00±0.00} & 0.00±0.00 & - & - \\
 & CLM Max & 1.00±0.00 & \textbf{1.00±0.00} & 0.00±0.00 & - & - \\
 & CLM Sum & 1.00±0.00 & \textbf{1.00±0.00} & 0.00±0.00 & - & - \\
 & GPS HS & 1.00±0.00 & \textbf{1.00±0.00} & \textbf{0.25±0.36} & - & - \\
 & GPS L & 1.00±0.00 & \textbf{1.00±0.00} & \textbf{0.25±0.36} & - & - \\
\bottomrule
\end{tabular}

    }
    \caption{Results for Llama 3 8B on TriviaQA}
\end{figure}
\begin{figure}[htbp]
    \centering
    \resizebox{\textwidth}{!}{  %
        \begin{tabular}{lllllll}
\toprule
 &  & Set Coverage & Abstention Rate & Non-Abs. Coverage & Set Size & Number of Samples \\
$\alpha$ & Method &  &  &  &  &  \\
\midrule
\multirow[t]{6}{*}{0.100000} & CLM FirstK & 1.00±0.00 & 1.00±0.00 & 0.00±0.00 & - & - \\
 & CLM FirstK (nr) & 1.00±0.00 & 1.00±0.00 & 0.00±0.00 & - & - \\
 & CLM Max & 1.00±0.00 & 1.00±0.00 & 0.00±0.00 & - & - \\
 & CLM Sum & 1.00±0.00 & 1.00±0.00 & 0.00±0.00 & - & - \\
 & GPS HS & 0.92±0.00 & \textbf{0.72±0.01} & \textbf{0.28±0.12} & \textbf{24.41±0.01} & \textbf{2.49±0.01} \\
 & GPS L & 0.99±0.00 & 0.97±0.00 & 0.16±0.20 & 24.86±0.04 & 2.82±0.07 \\
\bottomrule
\end{tabular}

    }
    \caption{Results for Phi 2 on TriviaQA}
\end{figure}

\end{document}